\documentclass[letterpaper, 10 pt, conference]{ieeeconf}

\usepackage{cite}
\usepackage{amsmath,amssymb,amsfonts}
\usepackage{algorithmic}
\usepackage{graphicx}
\usepackage{lipsum}
\usepackage{textcomp}
\usepackage{algorithm}
\usepackage{xcolor}
\usepackage{url}
\usepackage{booktabs} 
\usepackage{siunitx} 
\usepackage{caption} 
\usepackage[margin=1in]{geometry}

\IEEEoverridecommandlockouts                              

\begin{document}

\title{A YOLO-Based Semi-Automated Labeling Approach to Improve Fault Detection Efficiency in Railroad Videos}


\author{Dylan Lester$^{1}$, James Gao$^{1}$, Samuel Sutphin$^{1}$, Pingping Zhu$^{1*}$, Husnu S. Narman$^{2}$, Ammar Alzarrad$^{3}$
\thanks{This research was supported by the U.S. Army Engineer Research and Development Center (ERDC), grant \#W912HZ249C006.}
		\thanks{$^{1}$Dylan Lester, James Gao, Samuel Sutphin, and Pingping Zhu are with the Department of Computer Sciences and Electrical Engineering (CSEE), and Perception Intelligence Networks Group (PiNG),  Marshall University, Huntington, WV 25755, USA (lester299@marshall.edu, gao32@marshall.edu, sutphin54@marshall.edu, zhup@marshall.edu).}
		\thanks{$^{2}$Prof. Husnu S. Narman is with the Department of Computer Sciences and Electrical Engineering (CSEE), Marshall University, Huntington, WV 25755, USA (narman@marshall.edu)}
		\thanks{$^{3}$Prof. Ammar Alzarrad is with the Department of Civil Engineering (CE), Marshall University, Huntington, WV 25755, USA (alzarrad@marshall.edu).}
		\thanks{All correspondences should be sent to Prof. Pingping Zhu.}}



\maketitle

\begin{abstract}
Manual labeling for large-scale image and video datasets is often time-intensive, error-prone, and costly, posing a significant barrier to efficient machine-learning workflows in fault detection from railroad videos. This study introduces a semi-automated labeling method that utilizes a pre-trained You Only Look Once (YOLO) model to streamline the labeling process and enhance fault detection accuracy in railroad videos. By initiating the process with a small set of manually labeled data, our approach iteratively trains the YOLO model, using each cycle’s output to improve model accuracy and progressively reduce the need for human intervention.

To facilitate easy correction of model predictions, we developed a system to export YOLO’s detection data as an editable text file, enabling rapid adjustments when detections require refinement. This approach decreases labeling time from an average of 2–4 minutes per image to 30 seconds–2 minutes, effectively minimizing labor costs and labeling errors. Unlike costly AI-based labeling solutions on paid platforms, our method provides a cost-effective alternative for researchers and practitioners handling large datasets in fault detection and other detection-based machine learning applications.
\end{abstract}


\section{Introduction}\label{sec:intro}
Detection-based models such as YOLO have improved rapidly and are becoming increasingly accurate. However, training these models can be a time-consuming and labor-intensive process, with room for human error during the data preparation and training phases. While AI-assisted features exist to expedite training, they often come at a high cost, making them less accessible to researchers with limited resources. Finding methods to incorporate assisted labeling has shown to drastically improve accuracy , with Gregorio et al. seeing a 15\% increase over manual labeling methods\cite{degregorio2019semiautomatic}.  In this study, we propose a method that offers an effective and cost-efficient alternative to mainstream AI-assisted features. Specifically, we applied this method to detect faults within railroad systems, focusing on insufficient ballast—missing gravel between railroad tracks—and plant overgrowth. These faults can disrupt railroad traffic and pose safety risks.

Railroad fault detection has been extensively studied in the literature and continues to evolve with advances in AI technology. Railroad systems can fail for various different reasons. However, failure stems from the break down of the tracks. \cite{sener2022fault}. Most railroad fault detection processes have transitioned from manual to automated systems \cite{darm2024Automated, rivero2024application}, significantly enhancing efficiency and reliability. However, much of the existing research focuses on detecting cracks or structural issues in the rails themselves. Detecting insufficient ballast and plant overgrowth presents unique challenges due to the complexity of these faults and their subtle visual characteristics. During the initial stages of training a model to detect these specific faults, we encountered significant difficulties related to the labeling and training processes, which were both lengthy and demanding. To address these challenges, we developed an algorithm designed to improve model accuracy while substantially reducing the time required for training.To show improved accuracy this paper will compare a training set with only human labeled images with the same amount of images with the new algorithm we have developed. This paper outlines our approach, evaluates its performance, and discusses its implications for railroad fault detection and beyond. 

The remainder of this paper is organized as follows: Section \ref{sec:problem}, which discusses the complexities of developing the algorithm. Section \ref{sec:extLabels} covers the process of extracting and modifying the labels of the detected images. Section \ref{sec:Algo} is an in depth covering of the algorithm used. Section \ref{sec:result} analyzes the results of the algorithm used in training various models. Finally, Section \ref{sec:conclusion} has the final remarks.

\section{Problem Formulation}\label{sec:problem}

YOLO is a real-time object detecting model that classifies multiple objects through a single pass of a convolution neural network. When training a YOLO model, the availability of a large and diverse dataset is essential for achieving high accuracy and robust performance. Generally, the more extensive and representative the dataset used during training, the better the model's ability to generalize across various scenarios and detect objects accurately. For instance, in \cite{talaat2023,liu2024research}, over 10,000 images were employed to train a YOLO model effectively, demonstrating the scale required for successful model development.

However, preparing such a dataset presents significant challenges, particularly due to the need for image labeling. Each image must be annotated with bounding boxes and class labels to define the objects within them, a process that is both time-consuming and resource-intensive. Manual labeling can also be prone to errors, as it depends on human interpretation, which may vary among annotators. Inconsistent or inaccurate labeling can introduce noise into the training data, adversely affecting the performance metrics of the YOLO model, such as the F1-score and mean Average Precision (mAP).

The complexity of the objects being detected adds another layer of difficulty. In cases where object definitions are subjective or ambiguous, such as detecting insufficient ballast in a railroad dataset, labeling becomes even more challenging. Ambiguity in defining what constitutes "insufficient ballast" can lead to inconsistent annotations, further complicating the training process.

To address these issues, assisted labeling techniques can play a crucial role. By leveraging algorithms or semi-automated tools to assist in the labeling process, it becomes possible to reduce human error and increase consistency in annotations. Assisted labeling not only accelerates the dataset preparation process but also improves the overall quality of the labeled data.

In our lab, we encountered these challenges firsthand while training multiple YOLO models. The models often failed to achieve satisfactory F1-scores or mAP values, prompting us to explore alternative training techniques. We identified that developing a more efficient algorithm for assisted labeling could significantly enhance the training process. By ensuring that images are accurately labeled from the outset, the algorithm allows the model to extract better features and achieve superior performance, even with smaller datasets.

For our circumstances we chose to use the YOLOv8 model, as it is known to do well with rail oriented detection\cite{10477344}.However, this approach has broader implications beyond our specific use case. Developing and refining effective assisted labeling algorithms can improve model training across a wide range of fields, enabling researchers and practitioners to achieve better results with fewer resources. By addressing the bottlenecks in dataset preparation, such advancements can facilitate the application of YOLO models to diverse projects, from industrial inspections to real-time object detection systems.

\section{Extracting Labels}\label{sec:extLabels}
When a trained model is used to detect, it will read an image file from a given directory and analyze it, after it will output a label file. Labeled images are critical in training a model as it allows the model to learn what an object is. However, the images produced by our trained model must have their label files modified so that it could be edited within a labeling software.  We wrote a code to output an annotation file that had coordinates for each bounding box that was detected in the given image. Inside the annotation file there are values for the x center point, y center point, width, height, and class id. Another modification we had to make was creating a label map for the detected data so that it could be read by labeling software. The label map file contains the class id number and the class name. The class name is what will be read by labeling software. Using the trained model to detect objects in unlabeled images takes less than 5 seconds to label 100 images.

Once the images are uploaded into our labeling software, any necessary adjustments can be easily made. These modifications are automatically saved in a YOLO-readable format, ensuring seamless integration with YOLO detection frameworks. The label extraction process is executed during each detection of unlabeled images, making the overall workflow highly efficient.

\begin{figure}[h!]
    \centering
    \includegraphics[width=.8\linewidth]{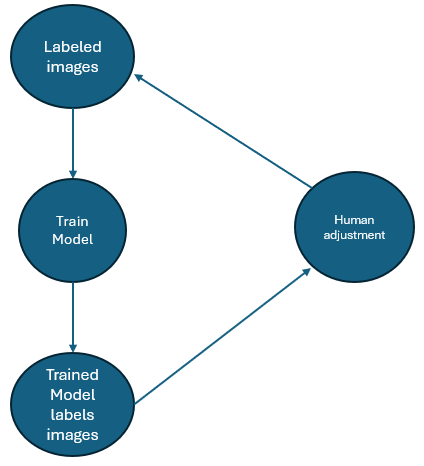}
    \caption{A diagram of the algorithm.}
    \label{fig:algo}
\end{figure}

\section{Algorithm}\label{sec:Algo}
Once a data set is acquired, a small subset of images from the collection will need to be labeled manually by a human annotator. This initial step is crucial when training a base YOLO (You Only Look Once) model, as it provides the foundational labeled data required to start the learning process. Labeled data serves as the ground truth for the model to understand the relationships between the input images and their corresponding outputs, such as bounding boxes and class labels.

After the images are labeled, the YOLO model is trained using this annotated subset. During this phase, the model learns to recognize patterns, extract features, and predict labels for objects within the training images. Once the training is complete, the model is deployed to label the next batch of images within the data set. This labeling process is done incrementally, ensuring that the model progressively refines its performance on increasingly larger data sets. By iteratively improving the model's predictions, less manual adjustment is required in subsequent cycles, ultimately enhancing efficiency.

For each iteration, the model's predictions on the new batch of images are reviewed and corrected using labeling software. These corrected labels are then incorporated into the existing labeled data set, expanding the training set and providing the model with additional examples for refinement. The updated labeled data set is used to retrain the YOLO model, further improving its accuracy and robustness.

As the data set grows and the model is exposed to a more diverse range of samples, its detection capabilities improve significantly. This progressive enhancement is crucial for deep learning models like YOLO, which thrive on large, well-annotated data sets. The incremental training approach not only reduces the time needed for manual review but also ensures that the model becomes better equipped to handle edge cases and outliers in the data.

Fig.\ref{fig:algo} is a visual display of how the algorithm functions. Algorithm \ref{alg:asstlabeling} goes into depth of how the algorithm functions.

\begin{algorithm} [h!]
\caption{Assisted Labeling}
\label{alg:asstlabeling}
\begin{algorithmic}[1]
\STATE \textbf{Procedure:} Manually label a number of images from the data set.
\STATE Train the YOLOv8 model on the labeled images.
\STATE Retrieve the best weight from the trained model.
\STATE Use the best weight to label images and edit the label files to be usable in a label editing software.
\STATE Add the detected images back into a labeling software and adjust the labels as needed.
\STATE Add the adjusted images into the labeled data, then train the model again.
\end{algorithmic}
\end{algorithm}

\section{Results}\label{sec:result}
\subsection{Dataset}
The dataset used in this study, is an open-source data set that consists of one railroad track \cite{rail_anomaly_detection}. A
total of 400 images are used for the training. The defects being detected are insufficient ballast,
and plants. The images of the railroad are at a top down angle which makes for a more accurate
viewing angle. 

To increase the total number of images that will be used in training, we applied augmentation to the data set. In a study on data augmentation for machine learning, Shorten, found that using data augmentation positively affected the training results \cite{shorten2019survey}. Table \ref{augmentations} outlines the augmentation settings that are applied to the data set.
\begin{table}[h]
    \centering
    \begin{tabular}{|c|c|}
        \hline
        flip & horizontal \\
        \hline
        rotation & between $-15^\circ$  and $+15^\circ$  \\
        \hline
       sheer & $\pm 10^\circ$ horizontal and $\pm 10^\circ$ vertical \\

        \hline
    \end{tabular}
    \caption{Data augmentations applied to the data sets.}
    \label{augmentations}
\end{table}
\begin{table}[h!]
    \centering
    \sisetup{table-format=4} 
    \begin{tabular}{S[table-format=3] S[table-format=3] S[table-format=4]}
        \toprule
        \textbf{Original Images} & \textbf{Augmented Images} & \textbf{Total Images} \\
        \midrule
        100 & 216 & 316 \\
        200 & 220 & 420 \\
        300 & 416 & 716 \\
        400 & 618 & 1018 \\
        \bottomrule
    \end{tabular}
    \caption{Summary of the image sets with augmentations.}
    \label{tab:image_set}
\end{table}

As seen in Table \ref{tab:image_set}, with the data augmentation we are able to increase the number of images used in the training by a significant amount. Pairing this with the algorithm in Algortihm \ref{alg:asstlabeling} and Fig. \ref{fig:algo}, we were able to optimize training a model for efficiency and accuracy.

\subsection{Result Evaluation}
YOLOv8 models can be objectively evaluated by using the mAP value, and F1-score \cite{wang2023bl}. mAP(Mean Average Precision) compares the bounding box with the models detection to return a score, and it is crucial that a accurate model has a higher mAP value. mAP value is calculated using Equation (\ref{eq:mAP}). The f1-score is used as a more holistic evaluation, computed by precision and recall. The F1-score is calculated using Equation (\ref{eq:f1}). 
\begin{equation}
\label{eq:mAP}
mAP = \frac{[\sum{P_A}]}{N}
\end{equation}
\begin{equation}
\label{eq:f1}
F1-score = 2*\frac{Precision*Recall}{Precision+Recall}
\end{equation}
There are two variations of mAP. mAP at 0.5 is the average precision at 0.5 threshold. mAP at 0.9 is the average precision at 0.9 threshold. In Equation (\ref{eq:mAP}), $N$ represents the number of classes, and $P_A$ is a numerical value of the area under a curve when the recall and precision are plotted \cite{talaat2023}. Recall is the models capability to identify false positive detections. It is crucial that an accurate model has a higher recall. Thus with precision and recall, the F1-score will be the models main evaluation method.

\subsection{Comparing Model Scores with Assisted Labeling}
\begin{figure}[!h]
 \centering
     \begin{minipage}{0.48\textwidth}
     \centering
        \includegraphics[width=\textwidth]{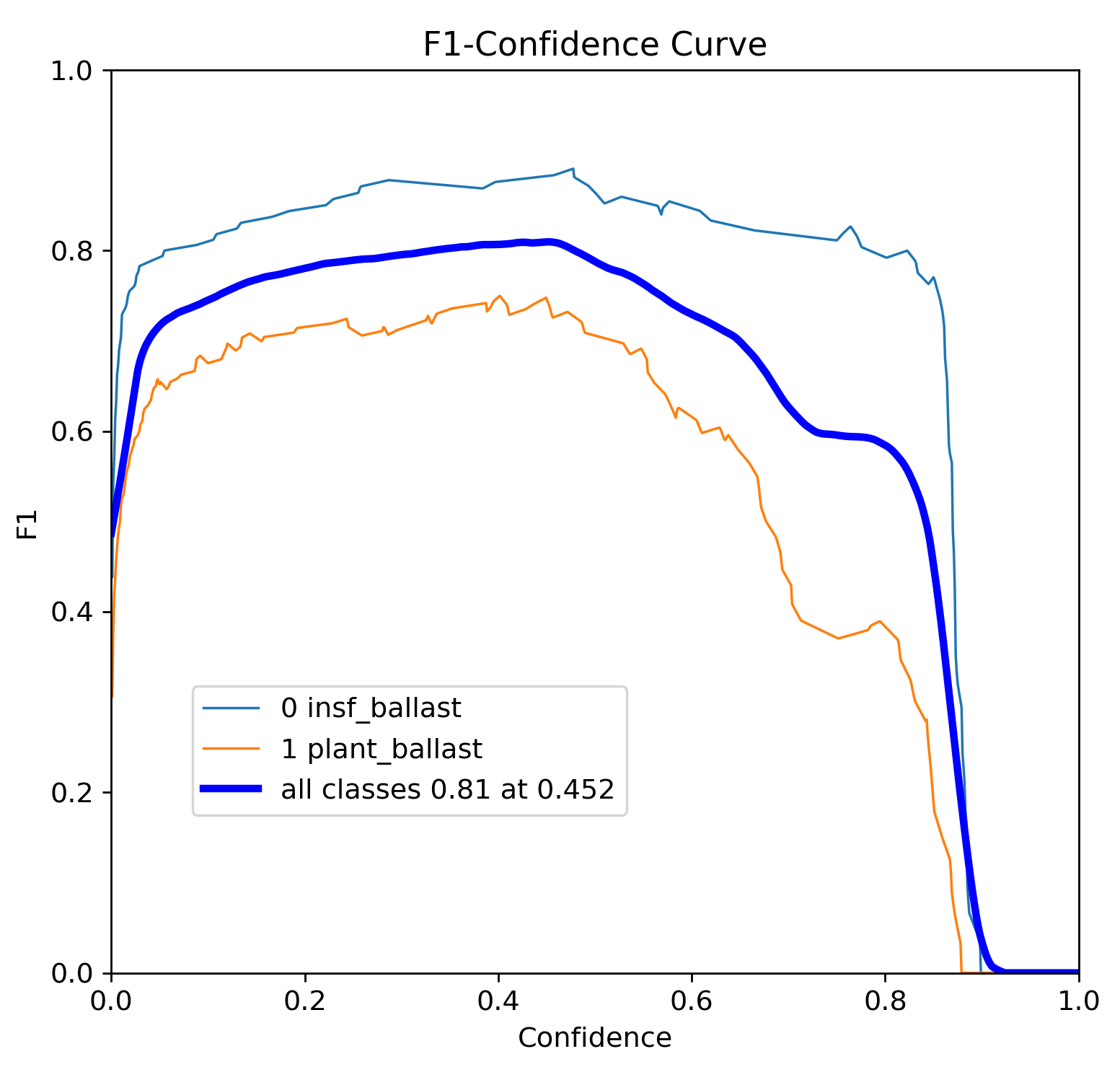}
        \caption{F1-score of the 100 image set.}
        \label{fig:100img}
    \end{minipage}
\begin{minipage}{0.48\textwidth}
        \centering
        \includegraphics[width=\textwidth]{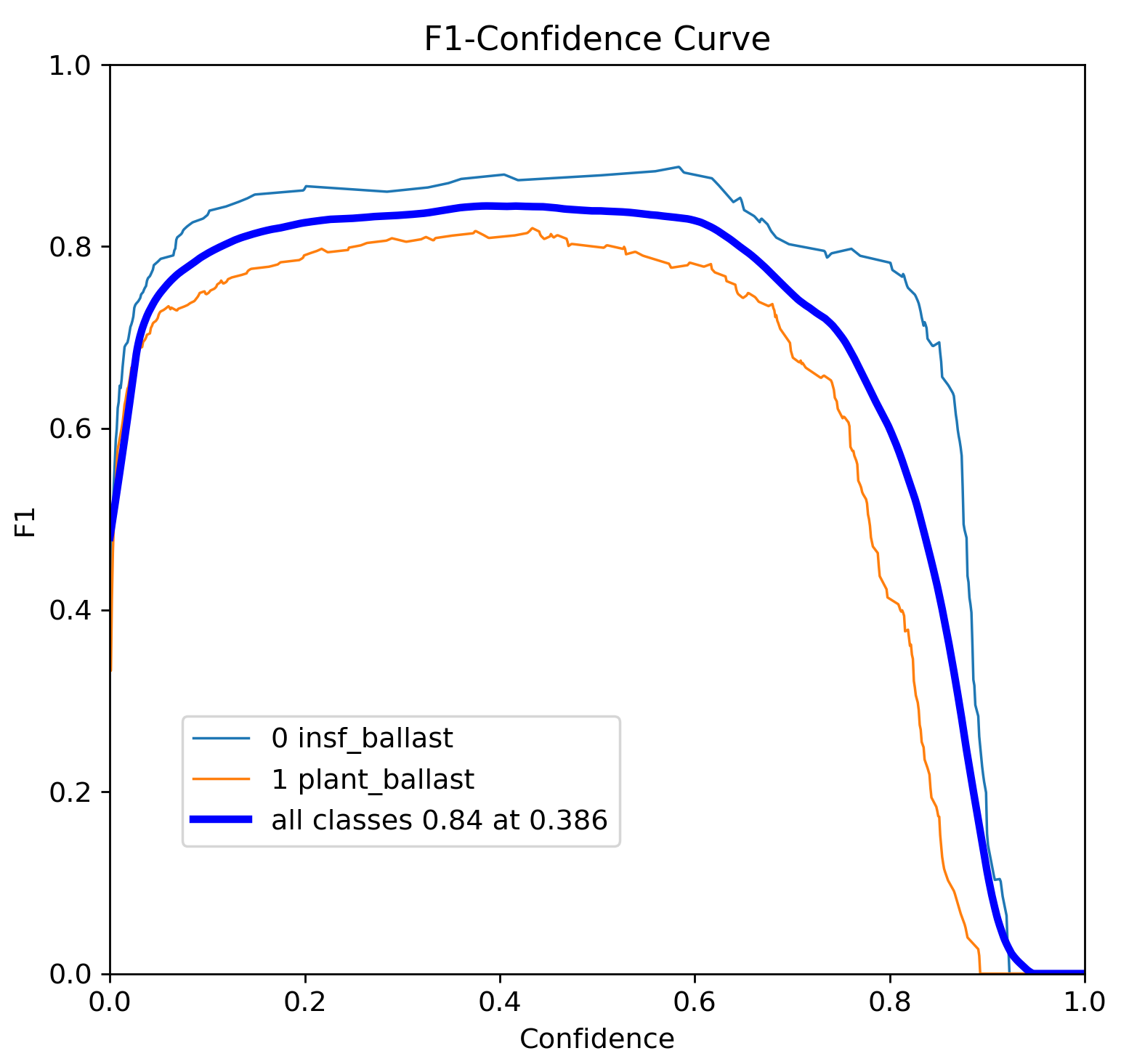}
        \caption{F1-score of the 200 image set.}
        \label{fig:200}
    \end{minipage}
\end{figure}
The 100 image set is the initial set with no assisted labeling. Due to the small data size and no assisted labeling the F1 score is not good as seen in Fig. \ref{fig:100img}. It can also be seen that the score is somewhat unstable.

\begin{figure} [h]
    \centering
    \begin{minipage}{0.48\textwidth}
         \centering
        \includegraphics[width=\textwidth]{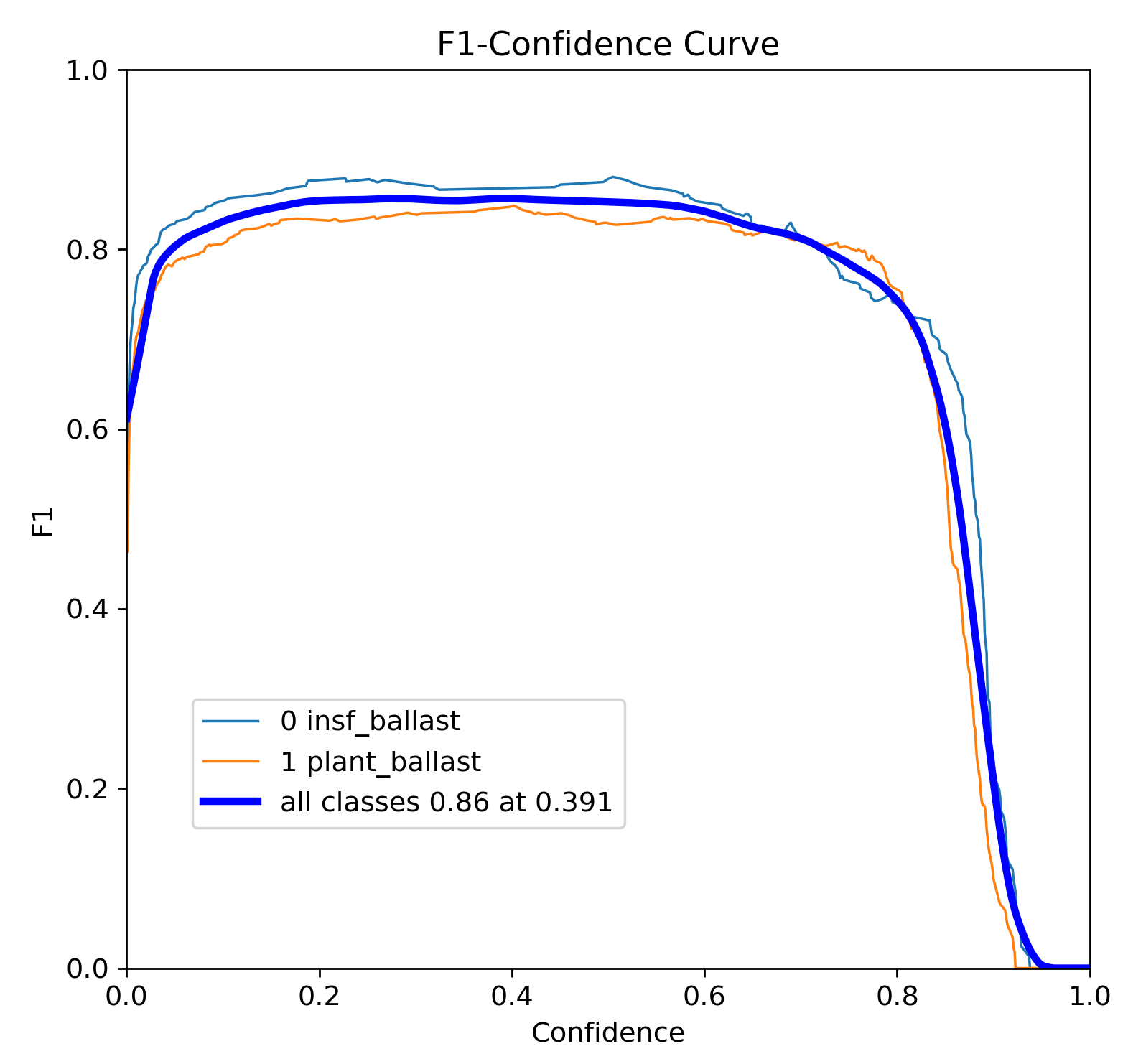}
        \caption{F1-score of the 300 image set.}
        \label{fig:300}
    \end{minipage}
    \begin{minipage}{0.48\textwidth}
        \centering
        \includegraphics[width=\textwidth]{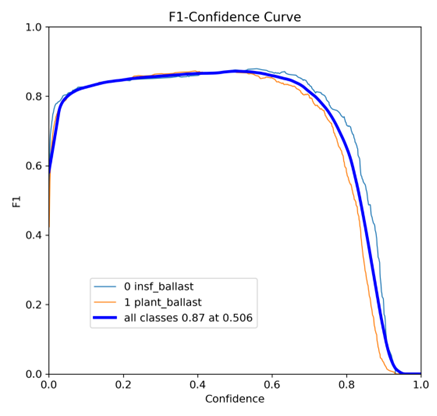}
        \caption{F1-score of the 400 image set.}
        \label{fig:400}
    \end{minipage}
\end{figure}

As shown in Fig. \ref{fig:200}, there is an initial increase in the score and stability of the model within the first implementation of assisted labeling. The 100 extra labeled images added from assisted labeling took significantly less time to add to the set as well. 
The same can be said for each additional 100 images, which can be seen in Fig. \ref{fig:300} and Fig. \ref{fig:400}. It can also be seen that the stability of the curve improves with each implementation of the assisted labeling highlighting that the model is getting more accurate with each incrimination of the algorithm. In total comparing the 100 image set with the 400 image set, the score increased from 0.81 to 0.87. With more data, the model would continue to improve. 
\subsection{Comparing Results With a Model With No Assisted Labeling}
For another comparison, we will compare the model with another model that has no assisted labeling. This model has a total of 400 images and has the same augmentations applied that are in Table \ref{augmentations}.
\begin{figure}[h!]
    \centering
     \begin{minipage}{0.48\textwidth}
         \centering
        \includegraphics[width=\textwidth]{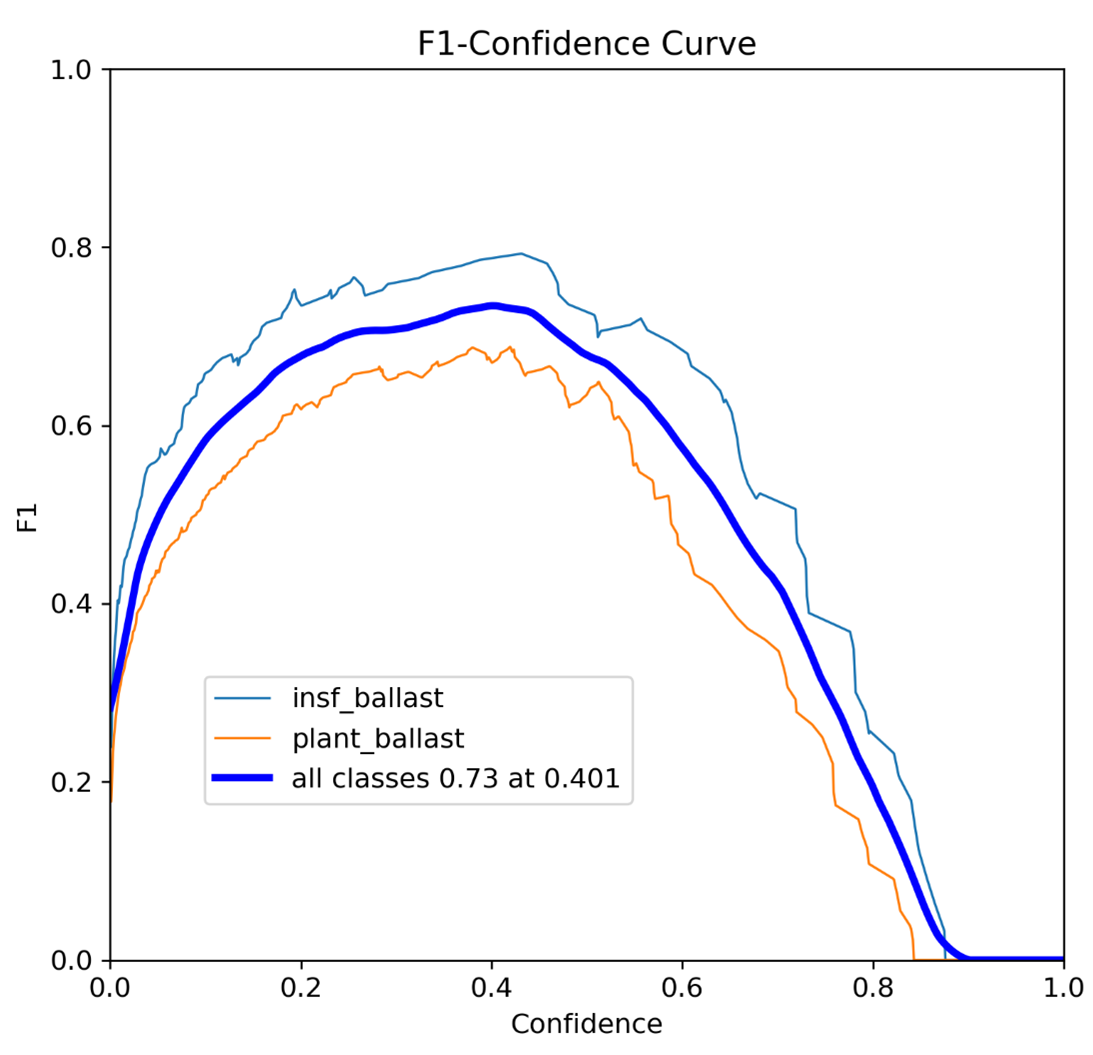}
        \caption{The unassisted model F1-score.}
        \label{fig:unassisted}
    \end{minipage}
    \begin{minipage}{0.48\textwidth}
         \centering
         \includegraphics[width=\textwidth]{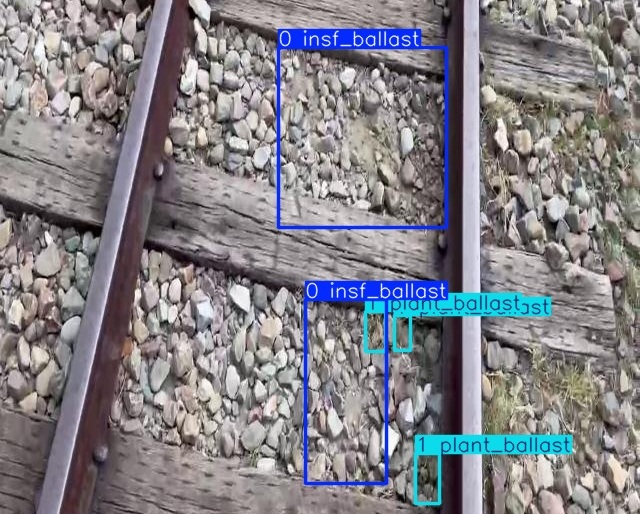}
        \caption{A sample detection from the most accurate model.}
        \label{fig:sample_detect}
    \end{minipage}
\end{figure}
As shown in Fig. \ref{fig:unassisted}, the model scored a 0.71 F1-score, and was notably unstable. The assisted model shows a drastic improvement in quality over the unassisted model. It must be noted that the unassisted model could have a fair amount of human error within the labeling process. However, this highlights another important feature of the algorithm which is the early implementation of assisted labeling which helps decrease human error in training from early stages. A difficult measurement to show the effectiveness of the assisted labeling is the time that it took to train the models. Manually labeling each image takes a significant amount of time. It took approximately 10 hours of labeling to fully label the unassisted model. To train the assisted model it took approximately 4-5 hours. This shows a significant time and labor reduction in the training process. Fig. \ref{fig:sample_detect} shows a sample output from using the 400 image model to detect an unlabeled image. For the purposes that we are using the model, this is an acceptable output. However, looking at Fig. \ref{fig:sample_detect}, shows the complexities of identifying insufficient ballast. It would be hard to come up with an exact definition of what would be insufficient ballast. However, the models output closely resembles the initial labeled data which implies an accurate detection. Table \ref{tab:f1_scores} is a summary of the models and their respective F1-scores.

\begin{table}[h!]
    \centering
    \caption{F1-Scores of the various models.}
    \sisetup{table-format=1.2} 
    \begin{tabular}{lc}
        \toprule
        \textbf{Data Set} & \textbf{F1-Score} \\
        \midrule
        100-image set (manually labeled)               & 0.81 \\
        200-image set (100 manual+100 assisted)              & 0.84 \\
        300-image set (100 manual+200 assisted)              & 0.86 \\
        400-image set (100 manual+300 assisted)               & 0.87 \\
        Baseline model (400 images manually labeled) & 0.89 \\
        \bottomrule
    \end{tabular}
    \label{tab:f1_scores}
\end{table}

\subsection{Effectiveness of Assisted Labeling Method}
The F1-score is a key metric for evaluating model effectiveness. As shown in Table \ref{tab:f1_scores}, the assisted labeling algorithm improves accuracy with each iteration as the dataset expands. This progressive enhancement occurs because the model refines its predictions with increased training data. A rising F1-score also indicates reduced human error, as mislabeled data can hinder model performance.

In most studies involving YOLO model training, including \cite{darm2024Automated} and \cite{rivero2024application}, manual labeling is the standard technique. While this method ensures high-quality annotations, it is time-consuming and costly. The proposed algorithm in this study incorporates a manual labeling component but significantly reduces the labor required compared to labeling an entire dataset.

Although the results demonstrate positive implications for training a model, the algorithm's performance depends on the quality of the initial labeled images. If the process begins with mislabeled data, the model's accuracy will suffer.

\section{Conclusion} \label{sec:conclusion}
Our research introduces a promising method for assisted labeling techniques that enhances both the speed and accuracy of training machine learning models. By employing our algorithm, we observed consistent improvements in F1-scores with each training iteration, alongside a progressively faster labeling process. This methodology facilitates more accurate and efficient detection of various railroad faults and can seamlessly integrate with any YOLO detection framework.

Future work will focus on implementing a confidence-level adjustment system, enabling the model to dynamically reduce the need for human intervention as its accuracy improves over iterations. This enhancement would significantly lower labor costs and further decrease training times. Additionally, we aim to refine our model for detecting insufficient ballast by acquiring a more specialized dataset tailored to our requirements, which we anticipate will lead to further improvements in performance metrics.


\bibliographystyle{IEEEtran}
\bibliography{references}

\end{document}